\documentclass[runningheads]{llncs}
\usepackage{graphicx}
\usepackage{bbding}
\usepackage{tikz}
\usepackage{comment}
\usepackage{amsmath,amssymb} 
\usepackage{color}

\usepackage[accsupp]{axessibility}  

\usepackage[width=122mm,left=12mm,paperwidth=146mm,height=193mm,top=12mm,paperheight=217mm]{geometry}

\begin{document}
\pagestyle{headings}
\mainmatter

\title{Faster-TAD: Towards Temporal Action Detection with Proposal Generation and Classification in a Unified Network} 


\titlerunning{Faster-TAD}
%
\author{Shimin Chen\thanks{These authors contributed equally to this work},
Chen Chen$^*$,
Wei Li,
Xunqiang Tao,
Yandong Guo}
%

\authorrunning{Chen et al.}
%
\institute{OPPO Research Institute\\
\email{\{chenshimin1,chenchen,liwei19\}@oppo.com}}

\maketitle

\begin{abstract}
Temporal action detection (TAD) aims to detect the semantic labels and boundaries of action instances in untrimmed videos. Current mainstream approaches are multi-step solutions, which fall short in efficiency and flexibility. In this paper, we propose a unified network for TAD, termed Faster-TAD, by re-purposing a Faster-RCNN like architecture. To tackle the unique difficulty in TAD, we make important improvements over the original framework. We propose a new Context-Adaptive Proposal Module and an innovative Fake-Proposal Generation Block. What's more, we use atomic action features to improve the performance. Faster-TAD simplifies the pipeline of TAD and gets remarkable performance on lots of benchmarks, i.e., ActivityNet-1.3 (40.01\% mAP), HACS Segments (38.39\% mAP), SoccerNet-Action Spotting (54.09\% mAP). It outperforms existing single-network detector by a large margin. 
\end{abstract}

\section{Introduction}
With the development of Internet, tons of videos are continuously generated in video content platforms, like YouTube, Netflix, Bilibili, etc. It shows that video understanding becomes more and more indispensable. Most videos are untrimmed in nature, thus temporal action detection is a fundamental task, which aims to detect start time, end time and semantic label of action instances. 

Current mainstream approaches~\cite{lin2018bsn,qing2021temporal,xu2020g} are multi-step solutions which achieve good performance. They include proposal generation, action classification, ensemble results of classifiers and proposal post-processing. However, they fall short in efficiency and flexibility, especially for videos with diverse semantic labels. In recent years, there are also some works focused on single network\cite{lin2021learning,xu2017r}, but they fail to yield comparable results as those of multi-step approaches.

To simplify the pipeline of TAD, we propose a novel single network with remarkable performance, dubbed Faster-TAD. Inspired by Faster-RCNN\cite{ren2015faster}, we jointly learn temporal proposal generation, action classification, and proposal refinement with multi-task loss, sharing information for end-to-end update.

We observe many challenges in temporal action detection. Firstly, to tackle the extreme duration variation of action instances, ranging from second to minute, boundary-based mechanism\cite{lin2019bmn} is adopted to generate proposals instead of the traditional anchor-based method. Secondly, proposal context is helpful for the recognition of the proposal label. To enhance semantic information for action instances, we bring an innovative Context-Adaptive Proposal Module for classification, in which we propose a new Proximity-Category Proposal Block to obtain context, a Self-Attention Block to construct relationships among proposals, and a Cross-Attention Block to learn relevant context in raw videos for proposals. Thirdly, we propose a new Fake-Proposal Block to make refinement module to be trained with various offsets relative to ground truth boundary. Last but not least, many complex human activities have long duration and consist of atomic actions, so action recognition models such as CSN\cite{tran2019video} are often adopted to extract the features of video clips as input for subsequent localization task. Nevertheless, action recognition model trained with limited input frames and complex human activity labels lacks atomic action information which may improve the recognition of boundaries and classes. To address this issue, we utilize atomic features trained on an atomic action dataset SEAL\cite{chen2022SEAL} as auxiliary features. 

To sum up, our contributions are as follow:
\begin{enumerate}
    \item We propose a unified network Faster-TAD for temporal action detection with the architecture which is similar to Faster-RCNN.
    \item In classification head, we propose a new Context-Adaptive Proposal Module, which consists of Proximity-Category Proposal Block, Self-Attention Block, and Cross-Attention Block. This Module greatly enhance semantic information for proposals.
    \item In proposal regression refinement, we propose a new Fake-Proposal Generation Block, with which we obtain more valid and diverse proposals for refinement.
    \item We find that feature representation is very important for this task, and demonstrate that atomic action features are helpful for complex activity detection.
\end{enumerate}

\section{Related Works}
\subsubsection{Temporal Action Proposal Generation.}
Temporal action detection task can be divided into temporal proposal generation and action classification. Approaches for action proposal generation can be grouped into two categories. The first method\cite{buch2017sst,gao2017turn,heilbron2016fast} is producing proposals with multi-scale anchors, which is inflexible and lack of boundary precision.  The second method generates proposals via locally locating temporal boundaries and globally evaluating the probability of potential action. BSN\cite{lin2018bsn} is the representative work. They later improved this work to BMN\cite{lin2019bmn}, which utilizes a Boundary-Matching confidence map to evaluate the probability of proposal globally. Boundary-Matching confidence map enumerates all possible combination of temporal locations, bringing promotion in both efficiency and effectiveness. Boundary-Matching confidence map can be called as an anchor set in extreme form. Recently, TCA-Net\cite{qing2021temporal} is proposed for temporal proposal refinement. G-TAD\cite{xu2020g} is produced to find effective video context.
\vspace{-0.5cm}
\subsubsection{Single Model Method.}
There are some works focused on single network, like Faster-RCNN-like works, Transformer-based works. Most Faster-RCNN-like works adopt pre-defined anchors to generate proposals. R-C3D\cite{xu2017r} encodes the video streams using a three-dimensional fully convolutional network, utilizing a streamline pipeline of Faster-RCNN. TAL-Net \cite{chao2018rethinking} improved receptive field alignment using a multi-scale architecture. GTAN \cite{long2019gaussian} introduced Gaussian kernels to dynamically optimize temporal scale of each action proposal. These works alleviate the problem of inflexible proposal generation to some extent, but still lack of boundary precision compared with boundary-based method and not taking full of semantic features of proposals. Recently, some Transformer-based works \cite{liu2021end,tan2021relaxed} are introduced, which encode proposal with transformer block. However, the problem of insensitivity to boundaries has not been effectively solved. To tackle these challenges of temporal action detection, we propose a single network named as Faster-TAD.

\begin{figure}[t]
\centering
\includegraphics[height=5cm,width=12cm]{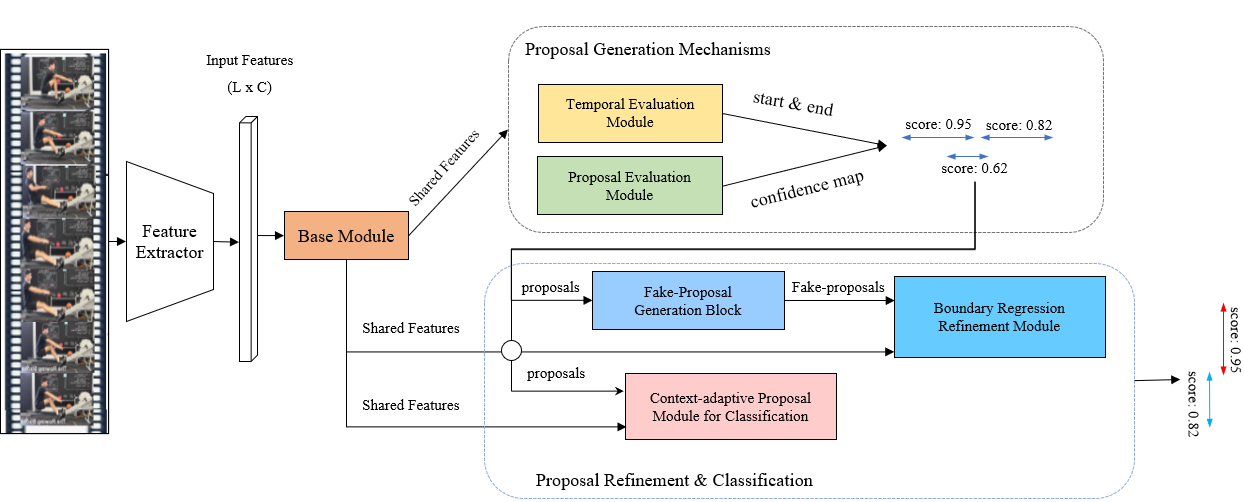}
\caption{Overview of our method. Given an untrimmed video, Faster-TAD can generate proposals and simultaneously (1) refine the boundary and (2) classify the proposal in a context-adaptive way. We construct our Faster-TAD with feature sequences extracted from raw video as inputs.}
\label{figure:faster-tad framework}
\end{figure}

\section{Our Approach}

\subsection{Overview of Framework}

As shown in Fig.~\ref{figure:faster-tad framework}, we propose a Faster-RCNN like network in temporal action detection, Faster-TAD. By jointing temporal proposal generation and action classification with multi-task loss and shared features, Faster-TAD simplifies the pipeline of TAD.

The input to our pipeline is a video sequence $X=\{x_t\}_{t=1}^T$, Following recent temporal action proposal generation methods \cite{buch2017sst,escorcia2016daps,lin2019bmn,xu2020g,zhao2017temporal}, we construct our Faster-TAD with feature sequences extracted from raw video as inputs by SwinTransformer\cite{liu2021swin} Extractor. The features of every $\tau$ consecutive frames are averaged and each set of the $\tau$ are named as a clip. In this way, input can be represented by $X\in\mathbb{R}^{C\times L}$, where $C$ is the feature dimension of each clip, and $L$ is the number of clips. 

We first process the feature sequences with a base module to extract shared features, which consists of a CNN Layer, a Relu Layer, and a GCNeXt\cite{xu2020g} Block. We then exert a Proposal Generation Mechanism to obtain most credible $K$ coarse proposals, where $K$ is 120. Proposals and shared features are further utilized to get more accurate boundaries by Boundary Regression Refinement Module\cite{qing2021temporal}. At the same time, shared features and proposals are employed to get the semantic labels of action instances with Context-Adaptive Proposal Module. We make some improvements to tackle the challenges in temporal action detection. 

\subsection{Proposal Generation Mechanism}
In anchor-based object detection tasks, anchors are generated in advance. The RPN Network\cite{ren2015faster} has two branches. Position Regression branch gets more accurate proposals, by regressing offset between anchors and ground truth regions. The Classification branch predicts positive score of anchors. Compared with object detection, a challenge of temporal action detection is the extreme duration variation of action instances. It is difficult to generate anchors to cover the receptive filed of all actions. So, we adopt Confidence-Matching mechanism\cite{lin2019bmn} to generate proposals instead of the traditional anchor-based method.

Proposal Generation Mechanism contains two branches, Temporal Evaluation Module(TEM) and Proposal Evaluation Module(PEM). We first adopt a Transformer Layer~\cite{vaswani2017attention} to get semantic information. Temporal Evaluation Module aims to evaluate the starting and ending probabilities for all temporal locations in untrimmed video. In Proposal Evaluation Module, we adopt SGAlign\cite{xu2020g} Block to generate Boundary-Matching (BM) confidence map, which aims to evaluate the probability of proposal globally. Boundary-Matching confidence map enumerates all possible combination of temporal locations, which can be called as an anchor set in extreme form. We use boundary probability sequences and BM confidence map to generate proposals during post processing.

\subsection{Context-Adaptive Proposal Module}

In object detection task, there is limited relation between objects in different spaces. However, in temporal action detection task, action instance is closely related to other actions in the same video. For example, if a semantic label of proposal is high jumping, the preceding action is likely to be running. The context greatly helps to classify the semantic label of proposals. We construct this module with Proximity-Category Proposal Block, Self-Attention Block, and Cross-Attention Block, learning useful context adaptively for each proposal.

\begin{figure}[t]
\centering
\includegraphics[scale=0.45]{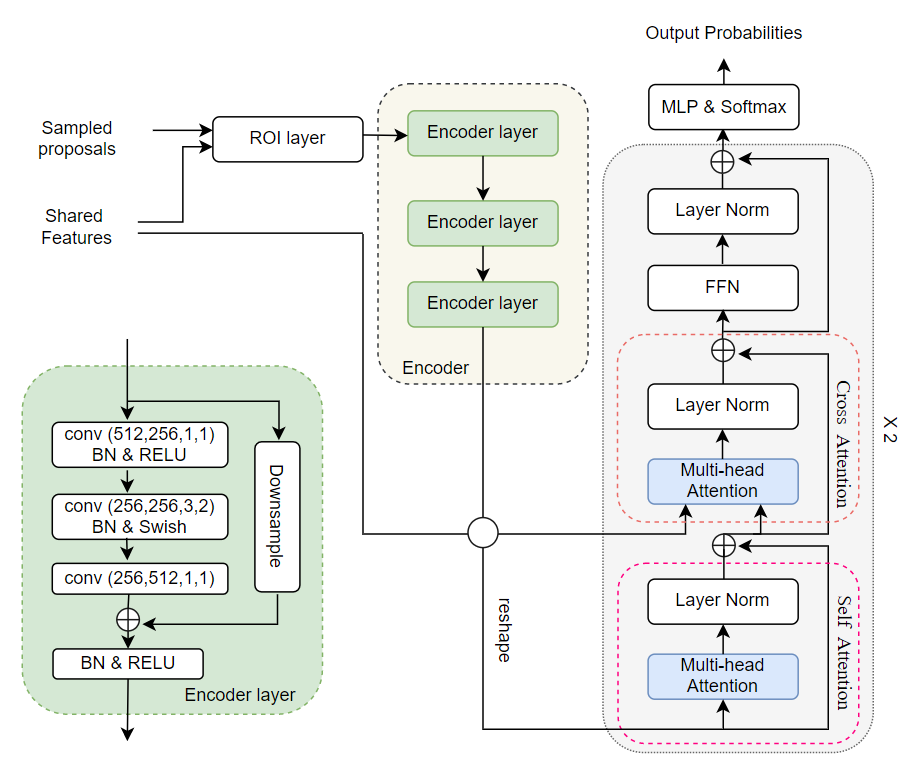}
\caption{Context-Adaptive Proposal Module. Proposal features are generated from proposal generation outputs and the shared features by a ROI layer. Then, encoder layer is followed to further encode the proposal representation. Finally, Self and Cross Attention block is applied to model the proposal semantic features.}
\label{fig:classification_head}
\end{figure}

Our Context-adaptive Proposal Module is illustrated in Fig.~\ref{fig:classification_head}. Proposal features are extracted from shared features by the ROI\cite{he2017mask} layer with sampled proposals. Compared with shared features ($X\in\mathbb{R}^{C\times L}$), Proposal features can be represented by $F_p\in\mathbb{R}^{N\times C\times T}$, where $N$ is the number of coarse sampled proposals, $C$ is the feature dimension of each clip, and $T$ is temporal resolution processed by ROI layer. The method to sample proposals are described in {``Proximity-Category Proposal Block’’} subsection. Encoder consists of three encoder layers to obtain deeper semantic features for each proposal. We employ Residual Block\cite{he2016deep} as the encoder layer. After three encoder layers, the temporal dimension of proposal features turn into $1/8 \times T$. Proposal features are afterwards flattened along the last two dimensions to a feature sequence ($P\in\mathbb{R}^{N\times \frac{T}8C})$, where $T$ is set to 16.

\begin{figure}
\centering
\includegraphics[scale=0.23]{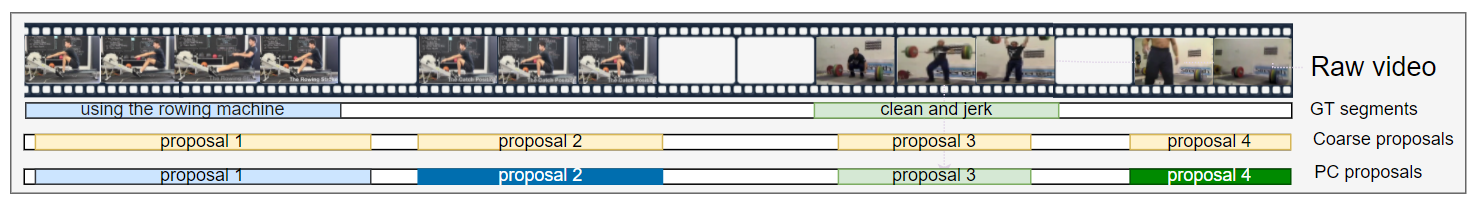}
\caption{Illustration of Proximity-Category proposals (PC proposals for short). The first row shows the ground truth segments. The second row is the output of our Proposal Generation Mechanism. The last row shows that the proposal with unsatisfied IoU will be assigned to a Proximity-Category according to its nearby ground truth segment. For example, proposal 2 has a label of {``using the rowing machine - proximity’’}.}
\label{proximity-category}
\end{figure}

\subsubsection{Proximity-Category Proposals Block}
Previous classification methods generally regard the proposal with IoU larger than a threshold as positive proposals, and other proposals as negative proposals. IoU is the matching score between the proposal and ground truth(GT).

In general, there are two methods to classify proposals\cite{liu2016ssd,redmon2016you,ren2015faster}. The first is to only reduce classification loss for positive proposals, the other is to add a negative category and assign all negative proposals to the negative category. However, negative proposals also contain many semantic information, and it isn't a sensible approach to classify all negative proposals into one category. As illustrated in Fig. \ref{proximity-category}, we propose a new proposal selection approach, called Proximity-Category Proposals Block. We define proposals with proximity category as Proximity-Category Proposals(PC Proposals). For example, in Fig. \ref{proximity-category}, proposal 2 is a PC Proposal of {``using the rowing machine - proximity’’}, and proposal 4 is a PC proposal of {``clean and jerk - proximity’’}. In this way, we expand the origin $M$ categories into $2M$ categories. We propose a high-low threshold method to set semantic labels for each proposal, according to the maximum IoU value between the proposal and ground true segments:
\begin{tiny}
\begin{equation}
\label{eq:sematic label}
{f_0}^{2M-1}(i)=\begin{cases}
f(i=idx)=1.0, f(i\neq idx)=0.0 &\text{if}\ IoU_{best}\geq{\tau}_1;\\[0.4cm]
\begin{split}
&f(i=idx)=\alpha IoU_{best},f(i=(idx+M))=1-\alpha IoU_{best}, \\
&f(i\neq {idx} \text{ and } i\neq {(idx+M)})=0.0
\end{split}
&\text{if}\ {\tau}_1>IoU_{best}\geq{{\tau}_2};\\[0.6cm]
f(i=(idy+M))=1.0,f(i\neq {(idy+M)})=0.0&\text{if}\ IoU_{best}<{\tau}_2;
\end{cases}
\end{equation}
\end{tiny}

\noindent where,
\begin{gather}
\label{eq:sematic label23}
idx=G(argmax(IoUs))\\
idy=G(argmin(Dists))
\end{gather}

As illustrated in Eq (\ref{eq:sematic label}), ${\tau}_1$ is the high threshold of $IoU$, ${\tau}_2$ is the low threshold. $M$ is the number of original categories, $i$ is ranging frame $0$ to $2M-1$ in Eq \ref{eq:sematic label}. ${IoUs}$ are the IoU values between proposal and the ground truth segments. ${IoU_{best}}$ is the max value of ${IoUs}$.  $idx$ and $idy$ are the index of label for proposal, $(idx+M)$ and $(idy+M)$ are index of Proximity-Category label for proposal. $\alpha$ is a hyperparameter. As illustrated in Eq (\ref{eq:sematic label23}), $G$ is a function that maps ground truth position to ground truth label index. ${Dists}$ stand for the center point distance values between proposal and the ground truth segments.

In order to make the ratio of positive proposals and PC Proposals close to 1:1, we select positive proposals first, and then select PC Proposals with the highest scores until $N$ coarse proposals are sampled.

\subsubsection{Self-Attention Block.}
We utilize Self-Attention\cite{vaswani2017attention} Block to capture relationships between proposals. Our decoder consists of 2 decoder blocks. At each block $l$, Query/Key/Value sequences are computed for proposal features from the representation ${P}^l$ encoder by the preceding block:
\begin{gather}
    {Q}^l={K}^l={V}^l={P}^l 
\end{gather}
Where $l$ is the block index. The multi-head variant of the self-attention computation is popularly used because of jointly attention to information from different representation sub-spaces. Multi-Head Attention\cite{vaswani2017attention} uses scaled dot-product attention, which is defined as:
\begin{gather}
    MultiHead-Attention(Q^l,K^l,V^l)=Concat({head_1}^l,...,{head_H}^l)W, \\
    {head_h}^l=softmax(\frac{{Q_h}^l\times {K_h^{Tl}}}{\sqrt{d}})V_h
\end{gather}
Where $H$ is the number of heads, $h$ is the head index, $W$ stands for parameter matrices, $d$ is the dimensionality of the hidden representations. 

We set the LayerNorm\cite{ba2016layer} after the Multi-Head Attention layer, followed with residual connection. We call this approach Middle-LN Transformer Layer, which is defined as:
\begin{equation}
    {P_{out}}^l=LN(MultiHead-Attention({Q}^l,{K}^l,{V}^l))+{Q}^l
\end{equation}

\subsubsection{Cross-Attention Block.}
We permute the shared features at the dimension of 0 and 1, which is encoded from raw video features by base module, and set the permuted features as key sequences and value sequences. It is represented by Eq (\ref{eq:cross-attn}). That is to say, Cross-Attention Block attend to learn the relationship between proposals and every clips in raw video. This block greatly increases the semantic information of proposals, bringing a large performance improvement.
\begin{equation}
    {K_{cross}}^l={V_{cross}}^l=permute(X),
    {Q_{cross}}^l={P_{out}}^l \label{eq:cross-attn}
\end{equation}
where $X\in\mathbb{R}^{C\times L}$ is shared features, $K_{cross}\in\mathbb{R}^{L\times C}$ is key sequences, $V_{cross}\in\mathbb{R}^{L\times C}$ is value sequences, $C$ is the feature dimension of each clip, and $L$ is the number of clips.

Like Self-Attention Block, we utilize Middle-LN Transformer Layer to learn relationships, which is defined as:
\begin{equation}
    {P_{cross}}^l=LN(MultiHead-Attention({Q_{cross}}^l,{K_{cross}}^l,{V_{cross}}^l)+{Q_{cross}}^l
\end{equation}
${P_{cross}}^l$ are further employed to get semantic features of proposals by a two-layer Feed-Forward Network(FFN)\cite{vaswani2017attention}, which is defined as:
\begin{equation}
    P^{l+1}=FFN({P_{cross}}^l)
\end{equation}

\subsection{Fake-Proposal Block}
\begin{figure}[t]
\centering
\includegraphics[scale=0.5]{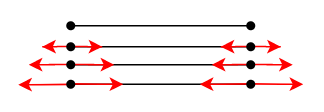}
\caption{Fake-Proposal Block. The red arrows indicate the different offsets based on the origin boundary of an ground truth segment. This block generates $[0,\pm\frac{1}{8},\pm\frac{1}{6},\pm\frac{1}{4}]$ seven modes to offset for each boundary of GT.}
\label{Fake-Proposal}
\end{figure}

 We propose a gt-based Fake-Proposal Block in Boundary Regression Refinement Module. We select the $\mu$ proposals with the highest confidence from $K$ coarse proposals as part of the proposals that will be refined. Then, we generate the rest $K-\mu$ fake proposals based on gt. As illustrated in Fig. \ref{Fake-Proposal}, we have $[0,\pm\frac{1}{8},\pm\frac{1}{6},\pm\frac{1}{4}]$ seven modes to change boundaries of GT. That is to say, each GT can generate 49 fake proposals. For each GT, we choose extension ways in order to generate fake proposals until $K-\mu$ fake proposals are generated, where $K$ and $\mu$ are set to 120 and 90.
 
The Fake-Proposal Block ensures that the refinement module can be trained with various boundary offsets relative to GT, which is beneficial for regression. On the other hand, this block feeds more valid proposals to the refinement module.

\subsection{Auxiliary-Features Block \label{Auxiliary-Features Block}}
\begin{figure}[t]
\centering
\includegraphics[scale=0.25]{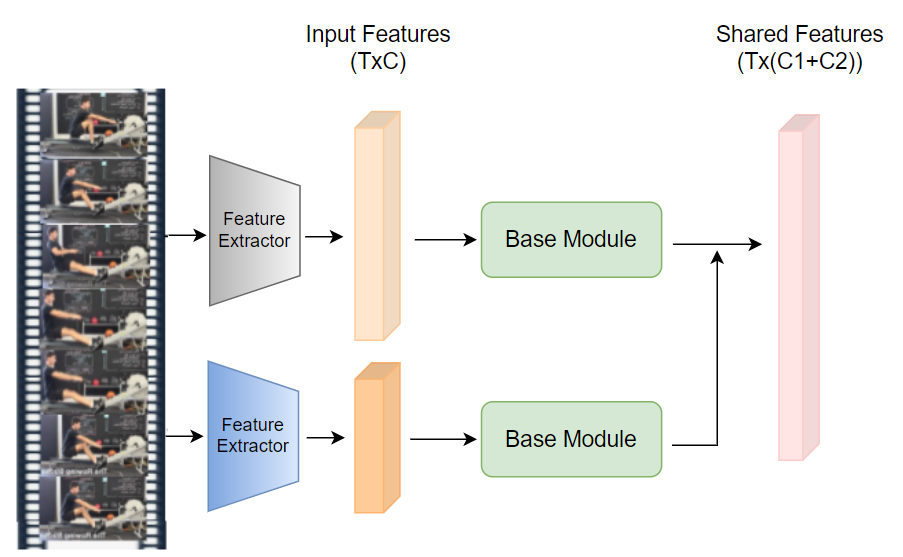}
\caption{Auxiliary-Feature Block. Two streams of features go through base module respectively. After that, they are combined along the temporal dimension. The rest of the network keeps the same.\label{Auxiliary-feature}}
\vspace{-0.5cm}
\end{figure}

 Many complex human activities have long duration and consist of atomic actions. Action recognition model Swin Transformer\cite{liu2021swin} is adopted to extract features of each clip as input for subsequent localization task. Nevertheless, action recognition model is trained with limited semantic information. To address this issue, we adopt atomic features as auxiliary features extracted by Slowfast\cite{feichtenhofer2019slowfast} trained on \textit{SEAL Clips}\cite{chen2022SEAL}. We designed a feature aggregation method named Auxiliary-Features Block to adapt to the two streams input. As shown in Fig. \ref{Auxiliary-feature}. Main and auxiliary features are combined in a simple way after going through two separate base modules. \textit{SEAL} \cite{chen2022SEAL} develop a novel large-scale video dataset of multi-grained spatio-temporally Action Localization. \textit{SEAL Clips} localizes atomic actions in space during a two-second clip.
 
\subsection{Training and Inference}
\subsubsection{Training.}
To jointly learn regressing temporal proposals and classifying action labels, an unified multi-task framework is exploited for optimization. The training details of Faster-TAD are introduced in this section. Once having both the coarse and refined prediction of temporal boundary, confidence score of positive proposal, and semantic label, we can optimize the model with the following objective function:
\begin{equation}
    L= {L_{C}}+{L_{R}}+\lambda{L_2}(\Theta)
\end{equation}
where $\lambda$ is hyper-parameter, set to 0.0001, ${L_2}(\Theta)$ is $L_2$ regularization loss, ${L_{C}}$ is loss of generating coarse proposals, and ${L_{R}}$ is loss of refining proposals and classifying labels. ${L_{C}}$ and ${L_{R}}$ are defied as:
\begin{gather}
    L_{C}= {L_{TEM}}^{BCls}+\lambda_1{L_{PEM}}^{BCls}+\lambda_2{L_{PEM}}^{Loc},\\
    L_{R}={L_{R}}^{Loc} +\gamma{L_{R}}^{Cls}
\end{gather}
where weight term $\lambda_1$, $\lambda_2$, and $\gamma$ are set to 1, 10, and 0.5, ${L_{TBR}}^{BCls}$ and ${L_{PEM}}^{BCls}$ are weighted binary logistic regression loss $L_{bl}$, adopted from BMN\cite{lin2019bmn}. we adopt $L_2$ loss for regression loss ${L_{PEM}}^{Loc}$. ${L_{R}}^{Loc}$ is a regression refined loss, adopted from TCA-Net\cite{qing2021temporal}.
 ${L_{R}}^{Cls}$ is softmax focal loss $\ell_{focal}$\cite{lin2017focal} between classification prediction and ground truth labels $y$:
\begin{equation}
    {L_R}^{Cls}=\frac{1}{2M}\sum_i^{2M} \ell_{focal}(\hat{y_i},{y_i})
\end{equation}
where $M$ is the number of original classification categories, $2M$ includes the number of original categories and proximity-categories.
\subsubsection{Inference.}
At inference time, top $K$ coarse proposals are proceeded to Context-Adaptive Proposal Module and Temporal Regression Refinement Module, predicting classification and regression score for each refined proposals. Following recent temporal action detection methods \cite{lin2019bmn,xu2020g,zhao2017temporal}, we construct predicted actions $\Psi=\left\{{{\psi}_k=((\hat{t}_{s,k},\hat{t}_{e,k}),\hat{c}_k,\hat{p}_k)}\right\}_{k=1}^K$, where $(\hat{t}_{s,k},\hat{t}_{e,k})$ refer to the predicted action boundaries, $\hat{c}_k$ is the predicted action category, and $\hat{p}_k$ is the fused confidence score of this prediction, $K$ is set to 120.

\section{Experiment}
\subsection{Datasets}
\vspace{-0.1cm}
\subsubsection{ActivityNet-1.3\cite{caba2015activitynet}}
 is a large-scale action understanding dataset for action recognition, temporal detection, proposal generation and dense captioning tasks. It contains 19,994 temporally annotated untrimmed videos with 200 action categories, which are divided into training, validation and testing sets by the ratio of 2:1:1.
\vspace{-0.38cm}
\subsubsection{HACS Segments\cite{zhao2019hacs}}
is a recently introduced dataset for temporal localization of human actions collected from web videos, therefore the results of many early methods on this dataset are not available. HACS Segments contains 139K action segments densely annotated in 50K untrimmed videos spanning 200 action categories.
\vspace{-0.38cm}
\subsubsection{SoccerNet-Action Spotting\cite{giancola2018soccernet}}
is a large-scale dataset for soccer video understanding. The dataset consists of 400 videos from soccer broadcast games for training, 100 videos for validation, and 50 separate games for test(challenge). In SoccerNet-Action Spotting, validation set is defined as test, and test set is called challenge. In action spotting task, 17 classes are annotated with a single timestamp, making annotations quite scattered in long videos. To convert the annotation format to traditional temporal action detection style, we take the annotated timestamp as the start moment and delay the timestamp by 4 seconds as the end moment. 

\subsection{Implementation Details}
\subsubsection{Features.}
For ActivityNet-1.3 and HACS Segments, we employ pre-extracted features as inputs. To fully demonstrate the effectiveness of our network, we use multiple features as input to conduct multiple experiments. The video features are extracted using Swin Transformer\cite{liu2021swin} trained on HACS Clips, which is called as Swin Feature in this paper. Auxiliary-Features are extracted by Slowfast\cite{feichtenhofer2019slowfast} trained on SEAL Clips\cite{chen2022SEAL}, which is defined as Slowfast-A Feature($A$ stands for atomic). We extract both features with windows of $size = 32$, $stride=32$. and we resize video features to 100 clips using linear interpolation. In addition, we also adopt TSP\cite{alwassel2021tsp} Feature trained on ActivityNet-1.3 as input. In training, we do not adopt any videos without actions. For SoccerNet Action Spotting, we train two feature extractors using Swin Transformer with 17 positive classes and 1 negative class. Two feature extractors have 3 and 6 second window sizes respectively to capture different context ranges. Features are combined using Auxiliary-Features Block mentioned in Chapter \ref{Auxiliary-Features Block}.
\subsubsection{Training and Inference.}
We train our model in a single network, with batch size of 64 on 8 gpus. The learning rate is $6\times{10}^{-4}$ for the first 3 epochs, and is reduced by 10 in epoch 3 and 7. We train the model with total 10 epochs. In inference, we apply Soft-NMS\cite{bodla2017soft} for post-processing, and select the top-M prediction for final evaluation. M is 120 for both ActivityNet-1.3 and Hacs Segments. We do not adjust hyper-parameters for HACS Segments, using the same hyper-parameters as those on ActivityNet-1.3.

\begin{table}[t]
    \centering
    \caption{Action detection results on validation set of ActivityNet-1.3, measured by mAP(\%) at different tIoU thresholds and the average mAP(\%). $S-N$ stands for Single-Network. $Ensemble \ of \ classifiers$ stands for ensemble video level classification results. $CUHK$ is from\cite{xiong2016cuhk}, $UntrimmedNet$ is from \cite{wang2017untrimmednets}. $Slowfast-A$ is extracted by Slowfast model trained on SEAL Clips\cite{chen2022SEAL}. {``-’’} indicates unknown results.}
    \label{tab:tab1}
    \scalebox{0.92}{
    \begin{tabular}{c|c|c|c|c|c|c|c}
    \hline
        Method & Feature & S-N & Ensemble of classifiers & 0.5 & 0.75 & 0.95 & Avg  \\\hline
        \multicolumn{8}{l}{Self-contained methods}\\\hline
        R-C3D\cite{xu2017r} & C3D & $\checkmark$ & $\times$ & 26.8 & - & - & - \\
        TAL-Net\cite{chao2018rethinking} & I3D & $\checkmark$ & $\times$ & 38.23 & 18.30 & 1.30 & 20.22  \\
        P-GCN\cite{zeng2019graph} & I3D & $\times$ & $\times$ & 42.90 & 28.14 & 2.47 & 26.99  \\
        TadTR\cite{liu2021end} & I3D & $\checkmark$ & $\times$ & 40.85 & 28.44 & 7.84 & 27.75 \\
        ContextLoc\cite{zhu2021enriching} & I3D &  $\checkmark$ & $\times$ & 51.24 & 31.40 & 2.83 & 30.59\\
        Lin et al.\cite{lin2021learning} & I3D & \checkmark & $\times$ & 52.4 & 35.3 & 6.5 & \textbf{34.4} \\
        \hline 
        \multicolumn{8}{l}{Ensemble Action-labels}\\\hline 
        P-GCN\cite{zeng2019graph} & I3D & $\times$ & UntrimmedNet & 48.26 & 33.16 & 3.27 & 31.11  \\ 
        BMN\cite{lin2019bmn} & TS & $\times$ & CUHK & 50.07 & 34.78 & 8.29 & 33.85  \\ 
        G-TAD\cite{xu2020g} & TS & $\times$ & CUHK & 50.36 & 34.60 & 9.02 & 34.09  \\
        TSP\cite{alwassel2021tsp} & TSP & $\times$ & CUHK & 51.26 & 37.12 & 9.29 & 35.81 \\
        BMN-CSA\cite{sridhar2021class} & TSP & $\times$ & CUHK & 52.64 & 37.75 & 7.94 & 36.25 \\
        TCANet[BMN]\cite{qing2021temporal} & Slowfast &$\times$ & CUHK & 54.33 & 39.13 &8.41 & 37.56\\ 
        PRN\cite{wang2021proposal} & CSN & $\times$ & PRN & 57.9 & - & - & \textbf{39.4} \\\hline
        \multicolumn{8}{l}{Our method}\\\hline
        Faster-TAD & TSP & \checkmark & $\times$ & 51.29 & 36.19 & 10.22 & 35.32\\
        Faster-TAD & TSP+Slowfast-A & \checkmark & $\times$ & 52.20 & 36.97 & 10.10 & 35.98 \\
        Faster-TAD & Swin & \checkmark & $\times$ & 57.39 & 39.97 & 10.48 & \textbf{39.09} \\
        Faster-TAD & Swin+Slowfast-A & \checkmark & $\times$ & 58.30 & 40.77&11.28 & \textbf{40.01} \\\hline
    \end{tabular}}
\end{table}

\subsection{Comparison with State-of-the-art Results}
\subsubsection{ActivityNet-1.3.}
Table \ref{tab:tab1} demonstrates the temporal action detection performance comparison on validation set of ActivityNet-1.3. Faster-TAD reports the highest average mAP results on this large-scale dataset. Our approach outperforms existing single-network detector by a large margin of 5.61\% mAP. Our single network outperforms these multi-step method by 0.61\% mAP, even multi-step detector using ensemble results of classifiers. 
\vspace{-0.1cm}
\subsubsection{HACS Segments.}
Table \ref{tab:tab2} compares Faster-TAD with state-of-the-art detectors on HACS Segments. Our approach obtains remarkable performance of 38.39\% average mAP, and outperforms existing single-network detector by a large maigin of 7.56\% mAP.
\vspace{-0.1cm}
\subsubsection{SoccerNet-Action Spotting.}
\cite{zhou2021feature} is the winner of SoccerNet-Action Spotting 2021. As shown in Table \ref{tab:tab3}, with Faster-TAD network, we reached an mAP of 54.09\% in tight test set, bringing a gain of 8.77\% mAP in shown set.

\begin{table}[t]
    \centering
    \caption{Action detection results on validation set of HACS-Segments, measured by mAP(\%) at different tIoU thresholds and the average mAP(\%). $S-N$ stands for Single-Network. $Ensemble \ of \ classifiers$ stands for ensemble video level classification results. $Slowfast-A$ is extracted by Slowfast model trained on SEAL Clips\cite{chen2022SEAL}. {``-’’} indicates unknown results. Results of BMN are from \cite{qing2021temporal}.}
    \label{tab:tab2}
    \scalebox{0.92}{
    \begin{tabular}{c|c|c|c|c|c|c|c}
    \hline
        Method & Feature & S-N & Ensemble of classifiers & 0.5 & 0.75 & 0.95 & Avg  \\\hline
        TadTR\cite{liu2021end} & I3D & $\checkmark$ & $\times$ & 45.16 & 30.70 & 11.78 & 30.83 \\
        G-TAD\cite{xu2020g}& I3D & $\times$ & - & 41.08 & 27.59 & 8.34 & 27.48  \\
        BMN\cite{lin2019bmn} & Slowfast & $\times$ & - & 52.49 & 36.38 & 10.37 & 35.76  \\ 
        TCANet[BMN]\cite{qing2021temporal} & Slowfast &$\times$ & - & 56.74 & 41.14 &12.15 & \textbf{39.77}\\ 
        \hline
        \multicolumn{8}{l}{Our method}\\\hline
        Faster-TAD & Swin & \checkmark & $\times$ & 54.13 & 37.10 & 12.02 & 36.92 \\
        Faster-TAD & Swin+Slowfast-A & \checkmark & $\times$ & 55.63 & 38.72 & 12.90 & \textbf{38.39} \\\hline
    \end{tabular}}
\end{table}

\begin{table}[t]
    \centering
    \caption{Action detection results on test set of SoccerNet-Action Spotting, measured by tight average-mAP introduced by \cite{giancola2018soccernet}. The test set is divided to shown part and unshown part according to the action visibility. We report the performance on Shown, Unshown and all test set. Validation set is defined as test, and test set is called challenge in SoccerNet.}\label{tab:tab3}
    \scalebox{1.0}{
    \begin{tabular}{c|c|c|c|c}
    \hline
        Method & Feature & Shown & Unshown & All \\\hline
        Zhao et al.\cite{zhou2021feature}(2021 top1) & BaiDu & 52.33&25.63&47.05 \\
        Faster-TAD(ours) & Swin(3s) & 56.91 & 24.38 & 50.34\\
        Faster-TAD(ours) & Swin(3s+6s) & 61.10 & 25.50 & 54.09 \\\hline
    \end{tabular}}
\end{table}

\begin{table}[t]
    \centering
    \caption{Ablation study of Single Network(SN), Proximity-Category(PC), Self-Attention(SA), Cross-Attention(CA), Fake-Proposal(FP) and Auxiliary-Features(AF) on ActivityNet-1.3 in terms of average mAP(\%). Each experiment is repeated five times, and we report the mean, standard deviation(std), and max values.}\label{tab:tab4}
    \begin{tabular}{c|c c c c c c |c|c|c}
    \hline
    Method  & SN & PC & SA & CA & FP & AF& mAP max & mAP mean & mAP std  \\\hline
    Faster-TAD & $\times$ & $\times$ & $\times$ & $\times$ & $\times$ & $\times$ & 38.02&37.97&0.048 \\
    Faster-TAD & \checkmark & $\times$ &$\times$&$\times$&$\times$&$\times$&38.21&38.12&0.064 \\
    Faster-TAD & \checkmark & \checkmark &$\times$&$\times$&$\times$&$\times$&38.68&38.57&0.098 \\
    Faster-TAD & \checkmark & \checkmark &$\checkmark$&$\times$&$\times$&$\times$&38.76&38.65&0.131 \\
    Faster-TAD & \checkmark & \checkmark &$\checkmark$&$\checkmark$&$\times$&$\times$&38.95&38.85&0.094 \\
    Faster-TAD & \checkmark & \checkmark &$\checkmark$&$\checkmark$&$\checkmark$&$\times$&39.09&38.99&0.068 \\
    Faster-TAD & \checkmark & \checkmark &$\checkmark$&$\checkmark$&$\checkmark$&$\checkmark$&40.01&39.88&0.123 \\\hline
    \end{tabular}
\end{table}

\subsection{Ablation Study}
As shown in Table \ref{tab:tab4}, in order to illustrate the effectiveness of each module, we performed five experiments for each configuration and obtained the maximum mAP, average mAP, and variance for each configuration. For the purpose of the stability of experiments, the average mAP is adopted below to prove the effectiveness of each module.
\subsubsection{Single-Network}
We use the video level classification results to replace the classification part of Faster-TAD, according to the general multi-step methods. Video level classification results are reached by using Swin Transformer model trained on HACS clips training set, obtaining 92.1\% top1 accuracy and 98.8\% top5 accuracy on HACS Clips validation set. The method is the same as video feature extraction of Faster-TAD. Table \ref{tab:tab3} shows that jointing temporal proposal generation and action classification with multi-task loss and shared feature can get better performance(+0.15\% mAP).
\subsubsection{Context-adaptive Proposal Module}
We optimize the model using the traditional method of reducing only the classification loss of positive proposals, and obtain 38.12\% mAP. On the other hand, we adopt the our Proximity-Category Proposal Module to reduce the classification loss, getting 38.57\% mAP with 0.45\% mAP improvement.

As shown in Table \ref{tab:tab4}, we adopt our Self-Attention Block to conduct relations among proposals, reaching 38.65\% mAP with 0.08\% gain.

By using our Cross-Attention Block to conduct relations between proposals and clips of raw video, we obtain a mAP gain of 0.2\%. It is demonstrated that our Context-adaptive Proposal Module effectively enhance semantic information for each proposal, bringing a gain of 0.73\% mAP.
\subsubsection{Fake-Proposal Block}
Table \ref{tab:tab4} reports the effectiveness of the Fake-Proposal Block, with which our network obtained 0.14\% mAP improvement.
\subsubsection{Auxiliary-Feature Block}
As shown in Table \ref{tab:tab4}, our Auxiliary-Feature has rich semantic information, with which our network obtained 0.89\% mAP improvement.
\section{Conclusions}
In this paper, We propose a novel network for temporal action detection in a single network, dubbed Faster-TAD. Faster-TAD includes Context-Adaptive Proposal Module to adaptively learn the semantic information of proposals by introducing attention mechanism across proposals to whole video and considering context as proximity-category proposals. Then the Fake Proposal based on the ground truth boundary with different offsets improves the Boundary Regression Module. Also, we found feature representation trained on atomic actions is very useful for complex activity detection. Our network can aggregate features with different semantic information and further improve the performance. Extensive experiments demonstrate that Faster-TAD outperforms existing single-network detector by a large margin on many benchmarks, obtaining state-of-the-art results on ActivityNet-1.3 and SoccerNet-Action Spotting.

\clearpage
%
%
\bibliographystyle{splncs04}
\bibliography{egbib}
\end{document}